\documentclass[letterpaper]{article} 
\usepackage[preprint]{aaai2027} 
\usepackage[hyphens]{url} 
\usepackage{graphicx} 
\usepackage{natbib} 
\usepackage{caption} 
\usepackage{booktabs}
\usepackage{amsmath}
\usepackage{amsfonts}
\usepackage{amssymb}
\usepackage{multirow}
\usepackage{xcolor}

\title{GeoNaviDiff: Transferring Disentangled Semantic Controls to Diffusion Features via Local Riemannian Navigation}
\author{
Tao Sun\textsuperscript{\rm 1},
Huanting Guo\textsuperscript{\rm 2},
Baao Xie\textsuperscript{\rm 1},
Zhou Fang\textsuperscript{\rm 3},
Qiuyu Chen\textsuperscript{\rm 1},
Hao Zheng\textsuperscript{\rm 1},
Wenjun Zeng\textsuperscript{\rm 1}
}
\affiliations{
\textsuperscript{\rm 1}Ningbo Key Laboratory of Spatial Intelligence and Digital Derivative, Ningbo Institute of Digital Twin, Eastern Institute of Technology, Ningbo, Zhejiang 315200, P.R. China\\
\textsuperscript{\rm 2}Beijing Institute of Technology, Beijing, P.R. China\\
\textsuperscript{\rm 3}Shaanxi Normal University
}

\begin{document}

\maketitle

\begin{abstract}

Disentangled real-image manipulation requires factor-specific semantic control while preserving image fidelity and non-target content. StyleSpace $\mathcal S$ provides interpretable controls but is limited by GAN inversion, whereas diffusion models offer stronger image realization without an explicit factorized control basis. We introduce GeoNaviDiff, an unsupervised cross-space framework that transfers disentangled StyleSpace interventions to local Riemannian trajectories in diffusion $h$-space. GeoNaviDiff constructs reconstruction-decoupled supervision by extracting a spherical Log-map tangent from a matched StyleGAN reconstruction–edit pair and parallel-transporting it to the real-image diffusion state. A shared, factor-conditioned Navigator predicts this tangent at inference, while norm-preserving geodesic realization and source-anchored rollout stabilize the reverse process without edited references or factor-specific models. Experiments across multiple datasets show that GeoNaviDiff outperforms the compared methods in fidelity, factor selectivity, edit locality, and trajectory smoothness.

\end{abstract}

\section{Introduction}
Disentangled Representation Learning (DRL)~\cite{bengio2013representationlearning} seeks to factorize complex observations into semantically meaningful variables, providing a foundation for interpretable and controllable generation. For representation manipulation, its value lies in selective intervention: a target factor, such as shape, color, or expression, should change while identity, structure, and other non-target factors remain stable. This factor-specific controllability is essential for fine-grained editing, yet it depends on both an interpretable representation and a generative model capable of faithfully realizing the prescribed change.

Early DRL-based manipulation methods primarily relied on the structured latent spaces of Variational Autoencoders (VAEs) and Generative Adversarial Networks (GANs)~\cite{NEURIPS2020_6fe43269,karras2019style}. In particular, StyleGAN exposes increasingly fine-grained semantic controls through its latent hierarchy, with channel-level coordinates in StyleSpace $\mathcal S$ supporting localized and interpretable interventions~\cite{karras2020analyzing,wu2021stylespace}. However, GAN-based real-image manipulation remains constrained by inversion fidelity and the fixed generator prior~\cite{tov2021designing,bobkov2024stylefeatureeditor,liu2023delving}. Diffusion models offer a stronger generative prior and high-quality image reconstruction, but their internal representations do not explicitly expose a factor-indexed semantic basis. This creates a fundamental tension: structured GAN representations provide factor-specific semantic controls, whereas diffusion priors favor high-fidelity reconstruction and synthesis.

\begin{figure}[t]
    \centering
    \includegraphics[width=\columnwidth]{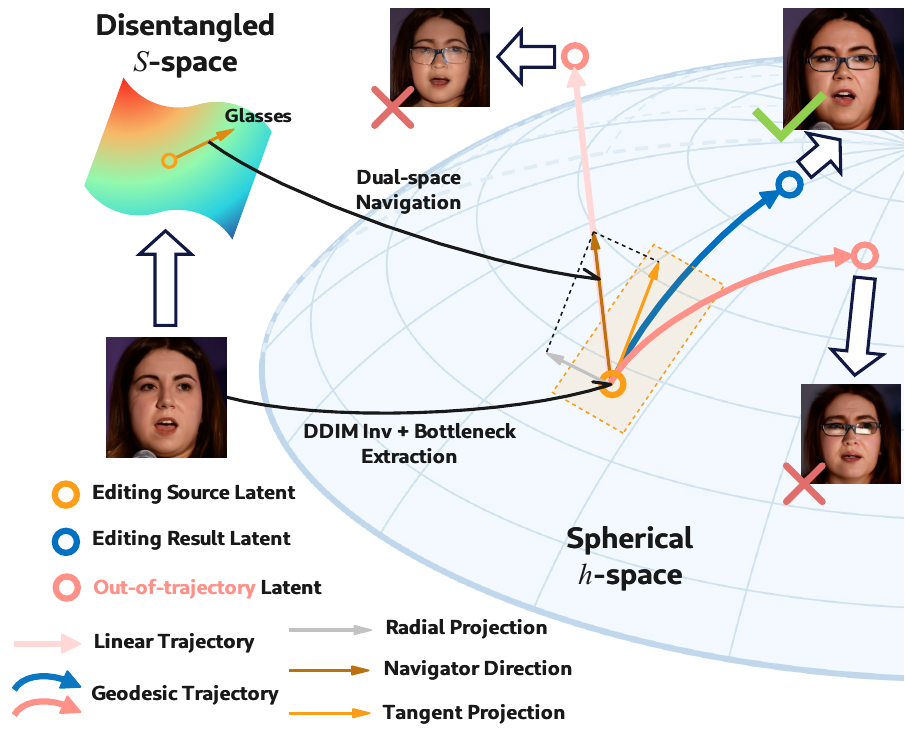}
    \caption{Conceptual comparison of Euclidean, arbitrary geodesic, and GeoNaviDiff trajectories in diffusion $h$-space.}
    \label{fig:fig1}
\end{figure}

To exploit the diffusion prior despite this missing factorized control space, existing conditional diffusion manipulation methods introduce semantic controls externally through text, attribute labels, or auxiliary guidance. Their controllable granularity, however, remains tied to predefined conditions and may cause unintended changes when those conditions are coarse or misaligned~\cite{parihar2024precisecontrolenhancing,hu2023selfguideddiffusion,sun2026context}. An intrinsic alternative is to manipulate the U-Net bottleneck feature space, known as $h$-space, which encodes rich high-level semantics without such external controls~\cite{kwon2022diffusionmodels,jeong2024trainingfreecontent}. Yet $h$-space does not expose these semantics as explicit, reusable control coordinates. Therefore, prior work trains a separate shift predictor for each prescribed semantic attribute and injects its output as an additive Euclidean residual~\cite{kwon2022diffusionmodels}. This direction-specific design prevents reuse across factors and ignores the geometry and feature statistics of $h$-space. As illustrated in Figure~\ref{fig:fig1}, its straight ambient path departs from the spherical shell of normalized bottleneck states and changes the bottleneck norm, disrupting its correlation with skip-feature statistics and producing visible artifacts~\cite{jeong2024trainingfreecontent}. Remaining on the hyperspherical shell alone is insufficient: an arbitrary geodesic may be geometrically valid while reaching a semantically incorrect endpoint. Therefore, stable disentangled manipulation requires a local tangent direction conditioned on an explicit semantic factor and adapted to the source image and diffusion timestep.

Motivated by this insight, we propose \textbf{GeoNaviDiff}, an unsupervised DRL framework for cross-space semantic transfer via \textbf{Geo}desic \textbf{Navi}gation in \textbf{Diff}usion $h$-space. Rather than disentangling diffusion $h$-space from scratch, GeoNaviDiff transfers unsupervised $\mathcal S$-space controls into it, enabling high-fidelity manipulation of individual semantic factors while preserving non-target content. A shared cross-space Navigator conditions on the $\mathcal S$-space displacement, source bottleneck state, and timestep to predict the corresponding local tangent direction in $h$-space. Geometry-aware learning and source-anchored inference then realize this prediction as a stable manifold-preserving diffusion trajectory without requiring an edited reference or a separate Navigator for each factor at inference. Our main contributions are:
\begin{enumerate}
	\item We introduce a cross-space paradigm that maps interpretable $\mathcal S$-space directions to stable semantic trajectories along the local hyperspherical geometry of diffusion $h$-space. This paradigm decouples semantic representation from image realization, with $\mathcal S$-space defining the semantics and the diffusion prior realizing them beyond GAN reconstruction fidelity.
	\item We develop a local Riemannian navigation mechanism with reconstruction-decoupled tangent supervision, norm-preserving geodesic realization, and source-anchored inference. This design aligns the transferred semantic direction with the real-image diffusion state while suppressing radial drift and cumulative rollout error.
	\item We conduct extensive evaluations across multiple datasets and diverse object categories. The results demonstrate that GeoNaviDiff generalizes robustly across object categories and achieves state-of-the-art performance across all reported metrics of image fidelity, representation disentanglement, edit locality, and trajectory smoothness.
\end{enumerate}

\section{Related Work}
\subsection{Disentanglement in VAEs and GANs}
Early disentangled representation learning primarily relied on variational autoencoders (VAEs), which separate generative factors by regularizing the latent distribution or encouraging factorized representations~\cite{burgess2018understandingdisentangling,kim2018disentangling,rolinek2019variational}. GAN-based approaches subsequently showed that interpretable semantic directions can be identified in pretrained generators through principal components, closed-form factorization, or unsupervised direction discovery~\cite{NEURIPS2020_6fe43269,shen2021closed,voynov2020unsupervised}. Moving beyond the dense directions of $\mathcal W$ and $\mathcal W^+$, StyleSpace Analysis exposed the channel-wise modulation coefficients of StyleGAN and showed that selected coordinates provide localized controls with stronger disentanglement and completeness under its evaluation protocols~\cite{karras2020analyzing,wu2021stylespace}. Fantastic Style Channels further selects representative and diverse channel directions from their generated image changes without using attribute labels~\cite{simsar2023fantastic}. These advances provide fine-grained semantic controls, but realizing them on real images remains constrained by GAN inversion and the fixed generator prior~\cite{xia2023ganinversion,tov2021designing}.

\subsection{Disentanglement--Fidelity Trade-off}
The tension between semantic editability and image fidelity is particularly evident in real-image manipulation. In StyleGAN, latent codes that remain close to the native distribution are generally more editable, whereas representations with greater freedom can reconstruct finer details at the cost of less stable manipulation; e4e, CLCAE, and SFE address different aspects of this trade-off through editability-aware inversion, cross-space guidance, or feature editing~\cite{tov2021designing,liu2023delving,bobkov2024stylefeatureeditor}. Diffusion-based DRL methods approach the complementary side of the problem by learning factorized representations while retaining the generative capacity of diffusion models: InfoDiffusion maximizes information in an auxiliary semantic latent variable, DisDiff learns factor-conditioned representations without supervision, and EncDiff uses cross-attention as an inductive bias for factor separation~\cite{wang2023infodiffusion,yang2023disdiffunsupervised,yang2024diffusioncrossattention}. These methods require training or modifying the diffusion model to learn a disentangled representation, rather than reusing an established semantic basis with a frozen denoiser. Inspired by these complementary advances, GeoNaviDiff uses $\mathcal S$-space to provide factor-specific semantic controls and the high-dimensional $h$-space of a pretrained U-Net denoiser as the realization space for their semantic shifts. This cross-space formulation combines the disentanglement of $\mathcal S$ with diffusion reconstruction fidelity without explicit conditioning or predefined $h$-space directions.

\subsection{Semantic Manipulation in Diffusion Models}
Diffusion models can be controlled through text, attribute labels, or auxiliary guidance, but the resulting editing granularity remains tied to the conditioning interface and may cause unintended changes when the condition is coarse or misaligned~\cite{parihar2024precisecontrolenhancing,hu2023selfguideddiffusion,sun2026context}. Intrinsic feature manipulation provides an alternative: Asyrp identifies the U-Net bottleneck $h$-space as semantically editable and learns a direction-specific predictor whose output is injected as an additive Euclidean residual~\cite{kwon2022diffusionmodels}, while subsequent work discovers semantic directions across diffusion timesteps without attribute supervision~\cite{park2023unsuperviseddiscovery}. InjectFusion further demonstrates that direct replacement or addition can disrupt bottleneck feature statistics, and improves content injection through norm-matched spherical interpolation and latent calibration~\cite{jeong2024trainingfreecontent}. These approaches establish the editability of diffusion features and the importance of preserving their geometry, but do not provide a shared factor-conditioned mechanism that maps an external disentangled control to a source- and timestep-adaptive tangent trajectory. GeoNaviDiff addresses this gap by learning a shared Navigator from $\mathcal S$-space displacements to local Riemannian navigation in diffusion $h$-space.

\section{Background}
\subsection{DDIM Sampler}
The Denoising Implicit Diffusion Model (DDIM) sampler \cite{song2020denoisingdiffusion} is a deterministic sampler that maintains consistency between forward and reverse process trajectories. This property enables it to map Gaussian noise $\hat x_T\sim \mathcal N(\mathbf 0, \mathbf I)$ back to generated images via smooth denoising trajectories:
\begin{equation}
\begin{aligned}
    \hat{x}_{t-1}
    ={}&
    \sqrt{\bar\alpha_{t-1}}
    \frac{
    \hat{x}_t-\sqrt{1-\bar\alpha_t}\,
    \boldsymbol{\epsilon}_\theta(\hat{x}_t,t,\mathbf c)
    }{\sqrt{\bar\alpha_t}}
    \\
    &+
    \sqrt{1-\bar\alpha_{t-1}-\sigma_t^2}\,
    \boldsymbol{\epsilon}_\theta(\hat{x}_t,t,\mathbf c)
    +\sigma_t\tilde\epsilon_t ,
\end{aligned}
    \label{eq:ddim_denoise}
\end{equation}
where $\sigma_t = \eta\sqrt{\frac{1-\bar\alpha_{t-1}}{1-\bar\alpha_{t}}}\cdot\sqrt{1-\frac{\bar\alpha_{t}}{\bar\alpha_{t-1}}}$, $\eta$ is a hyperparameter that controls the stochasticity of the sampling process. In the deterministic form (e.g., $\eta=0$), DDIM is expected to recover the same noise across all sampling steps as in the corresponding denoising steps, thereby achieving consistent inversion of the diffusion process. 

\subsection{Asymmetric Reverse Process}
Asymmetric Reverse Process (Asyrp) \cite{kwon2022diffusionmodels} enables image editing by injecting offset priors into the semantically editable $h$-space during the denoising process. Let $h_t$ be the original $h$ latent matrix corresponding to timestep $t$. From the unconditional form of Eq. \eqref{eq:ddim_denoise}:
\begin{equation}
\begin{aligned}
    P_t(\tilde x_t, t; h_t) &= \frac{\tilde{x}_t-\sqrt{1-\bar\alpha_t}\cdot\mathbf{\epsilon}_\theta(\tilde{x}_t,t; h_t)}{\sqrt{\bar\alpha_t}}, \\
    D_t(\tilde x_t, t; h_t) &= \sqrt{1-\bar\alpha_{t-1} - \sigma^2_t}\cdot\mathbf{\epsilon}_\theta(\tilde{x}_t,t; h_t).
\end{aligned}
\label{eq:asyrp_branches}
\end{equation}
Asyrp uses a simulated diffusion process via DDIM inversion to obtain the semantically manipulated $\tilde h_t$ and defines the following three-stage editing process. For compactness, let $P_t(h)=P_t(\tilde x_t,t;h)$, $D_t(h)=D_t(\tilde x_t,t;h)$, and denote the three timestep intervals by $\mathcal I_1=[t_\text{edit},T]$, $\mathcal I_2=[t_\text{boost},t_\text{edit})$, and $\mathcal I_3=[0,t_\text{boost})$:
\begin{equation}
    \tilde{x}_{t-1}
    =
    \begin{cases}
        \sqrt{\bar{\alpha}_{t-1}}\,P_t(\tilde{h}_t) + D_t(h_t),
        & t \in \mathcal{I}_1, \\[2pt]
        \sqrt{\bar{\alpha}_{t-1}}\,P_t(h_t) + D_t(h_t),
        & t \in \mathcal{I}_2, \\[2pt]
        \sqrt{\bar{\alpha}_{t-1}}\,P_t(h_t) + D_t(h_t)
        + \sigma_t \tilde{\epsilon}_t,
        & t \in \mathcal{I}_3.
    \end{cases}
\end{equation}
The editing process is performed within the interval $T \geq t \geq t_\text{edit}$. It injects the modified $\tilde h_t$ only into $P_t(\cdot)$ to prevent the injected semantics from being canceled out by $D_t(\cdot)$.

\section{Methodology}
\subsection{Overview}
\begin{figure*}[t!]
    \centering
    \includegraphics[width=\textwidth]{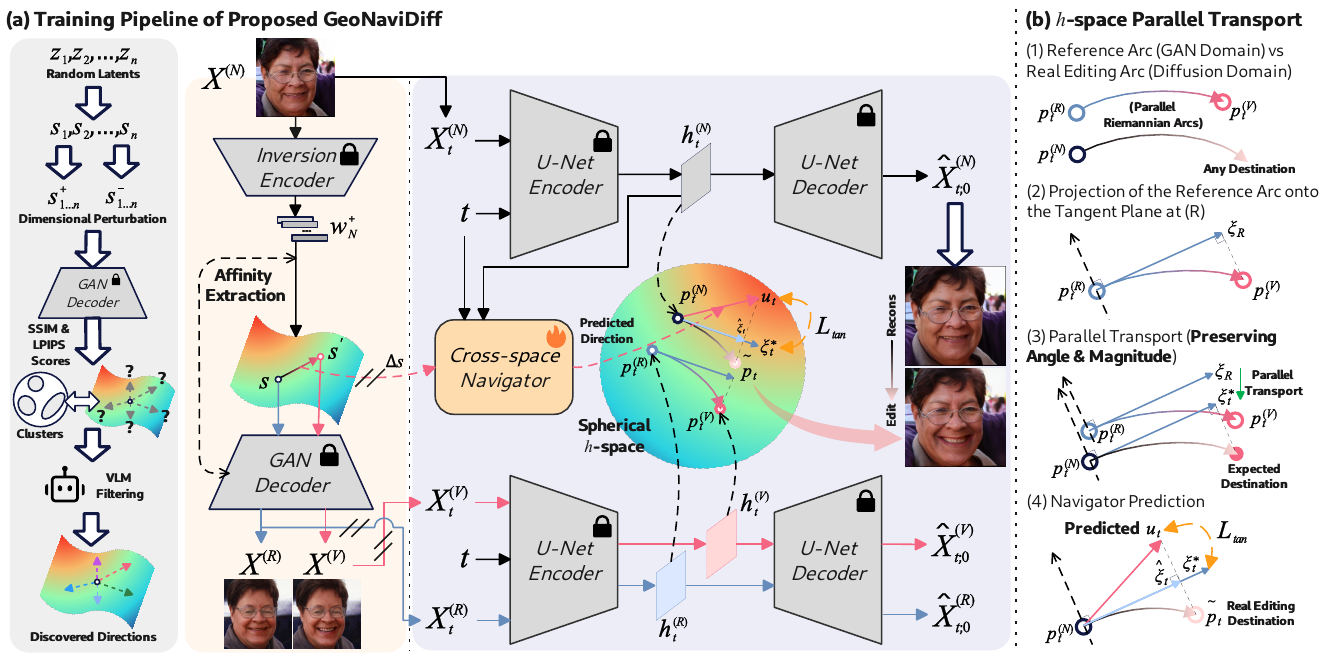}
    \caption{Training framework of GeoNaviDiff. (a) Unsupervised semantic discovery and disentangled control construction in \(\mathcal S\)-space, followed by cross-space navigation into diffusion \(h\)-space. (b) The four-step parallel-transport procedure is used to construct the tangent navigation target.}
    \label{fig:overall}
\end{figure*}

As shown in Figure~\ref{fig:overall}, GeoNaviDiff transfers unsupervised, factor-specific $\mathcal S$-space controls into diffusion $h$-space to enable high-fidelity, disentangled manipulation of real images. Given a source image and an intervention $\Delta s$, the StyleGAN branch constructs a matched reconstruction--edited reference pair, and a frozen U-Net extracts bottleneck states for the source and both references under shared noising. The Riemannian displacement between the reference states defines a semantic tangent, which is parallel-transported to the real-image source state to form the Navigator target. The Navigator predicts this target from $(\Delta s,t,h_t^{(N)})$; norm-preserving geodesic realization and source-anchored inference then turn the prediction into a stable diffusion trajectory.
\subsection{Disentangled Semantic Basis in $\mathcal S$-space}
GeoNaviDiff uses StyleGAN2's StyleSpace $\mathcal S$ as its factorized semantic basis. Given an inverted code $w^{+}=\{w_{\ell}\}_{\ell=1}^{L}$, the layer-wise affine transforms yield
$s=(A_1(w_1),\ldots,A_L(w_L))\in\mathcal S=\prod_{\ell=1}^{L}\mathbb{R}^{C_{\ell}}$,
where each coordinate modulates a synthesis channel~\cite{karras2020analyzing,wu2021stylespace}. Selected coordinates provide localized controls; the resulting displacement $\Delta s$ both conditions the Navigator and generates the matched reconstruction--edit pair used for tangent supervision.

Since $\mathcal S$-space contains thousands of channels with redundant effects, we adopt the unsupervised Fantastic Style Channels method~\cite{simsar2023fantastic} to discover representative and diverse candidate directions. It evaluates axis-aligned channel perturbations through their generated image changes and groups channels with similar effects into candidate clusters. As shown in Figure~\ref{fig:overall}(a), a pretrained vision-language model (VLM) filters these clusters by inspecting their traversal sequences to retain semantically meaningful directions; the same filtering can instead be performed manually. Let $\mathcal{I}=\{i_k\}_{k=1}^{K}$ denote the retained Fantastic Style Channels indices and $d_k=e_{i_k}$ the corresponding coordinate basis vector. During Navigator training, a sampled direction and strength specify $\Delta s=\alpha d_k$ and $s'=s+\Delta s$, where the sign and magnitude of $\alpha$ determine the traversal direction and extent, respectively.

For a real image $x$, a pretrained and frozen e4e inversion encoder~\cite{tov2021designing} produces $w^{+}$, and the fixed affine transforms above yield $s$. The frozen StyleGAN generator then produces the matched pair $X^{(R)}=G(s)$ and $X^{(V)}=G(s+\alpha d_k)$.
This pair shares the same inverted code and generator pathway and differs only in the prescribed $\mathcal S$-space intervention. It therefore externalizes the factor-specific change encoded by $\Delta s$ as a matched visual reference for constructing the diffusion-space navigation target in Sec. ~\ref{sec:method_navigation}. In GeoNaviDiff, $\mathcal{S}$ thus supplies both the factor-indexed control descriptor and its visual supervision.
\subsection{Cross-Space Riemannian Navigation}\label{sec:method_navigation}
Existing $h$-space editors commonly inject an ambient Euclidean residual, $\widetilde h_t=h_t+\lambda\Delta h_t$~\cite{kwon2022diffusionmodels}, ignoring bottleneck geometry and feature statistics:
\begin{equation}
    \|\widetilde h_t\|_F^2
    =
    \|h_t\|_F^2
    +2\lambda\langle h_t,\Delta h_t\rangle_F
    +\lambda^2\|\Delta h_t\|_F^2 ,
    \label{eq:euclidean_norm_drift}
\end{equation}
so Euclidean injection incurs radial leakage and, even for a tangent residual satisfying $\langle h_t,\Delta h_t\rangle_F=0$, increases the squared norm by $\lambda^2\|\Delta h_t\|_F^2$. This is consequential because bottleneck and skip-feature norms are correlated, and disrupting this relation produces visible artifacts~\cite{jeong2024trainingfreecontent}. Moreover, the linear interpolation between two distinct normalized states passes through the interior of the sphere rather than following its surface, producing an off-shell trajectory.

To the best of our knowledge, GeoNaviDiff is the first framework to perform local Riemannian navigation in diffusion $h$-space conditioned on disentangled $\mathcal S$-space factors. It replaces this geometry-agnostic residual with cross-space navigation from $\mathcal S$-space to diffusion $h$-space. Here, each factor is indexed by an unsupervised $\mathcal S$-space direction $d_k$, and its signed magnitude is specified by $\Delta s=\alpha d_k$. Rather than claiming that the complete $h$-space is globally spherical, we locally model normalized bottleneck states at each timestep on the unit hypersphere $\mathbb S^{d-1}$. The resulting navigation target is sample-dependent and timestep-dependent, and explicitly conditioned on $\Delta s$.

\paragraph{Matched semantic triple.}
As described in Sec.~4.2, we construct an unedited StyleGAN reconstruction $X^{(R)}$ and its edited reference $X^{(V)}$ for each real source image $X^{(N)}$. We apply the forward noising process to all three images using the same sampled timestep $t$ and noise realization $\epsilon$:
\begin{equation}
\left\{
\begin{aligned}
    X_t^{(N)} &= \sqrt{\bar\alpha_t}X^{(N)}
        + \sqrt{1-\bar\alpha_t}\epsilon,\\
    X_t^{(R)} &= \sqrt{\bar\alpha_t}X^{(R)}
        + \sqrt{1-\bar\alpha_t}\epsilon,\\
    X_t^{(V)} &= \sqrt{\bar\alpha_t}X^{(V)}
        + \sqrt{1-\bar\alpha_t}\epsilon .
\end{aligned}
\right.
\label{eq:matched_noising}
\end{equation}
Shared noising removes stochastic differences from the comparison. More importantly, $X^{(R)}$ and $X^{(V)}$ share the same inversion and synthesis path and differ only by $\Delta s$; their difference therefore isolates the prescribed semantic intervention, whereas $X^{(N)}$ supplies the real-image state encountered by the Navigator at deployment. With the frozen U-Net bottleneck extractor $H_\psi$, we obtain
\begin{equation}
    h_t^{(i)}=H_\psi(X_t^{(i)},t),
    \qquad i\in\{N,R,V\}.
    \label{eq:h_pair}
\end{equation}
Together, these three bottleneck states separate the StyleGAN semantic change from the real-to-reconstruction gap before transferring it into diffusion $h$-space.

\paragraph{Spherical normalization and semantic Log map.}
Let $r_i=\|h_t^{(i)}\|_F$. We normalize each bottleneck state as
\begin{equation}
    p_t^{(i)}=\frac{h_t^{(i)}}{r_i}\in\mathbb S^{d-1},
    \qquad i\in\{N,R,V\}.
    \label{eq:sphere_normalization}
\end{equation}
We first measure the semantic edit entirely within the matched reconstruction domain. Define
\begin{equation}
\begin{aligned}
    c_{RV} &= \langle p_t^{(R)},p_t^{(V)}\rangle_F,
    &\qquad
    s_{RV} &= \sqrt{1-c_{RV}^{\,2}},\\
    \theta_{RV} &= \operatorname{atan2}(s_{RV},c_{RV}),
    &\qquad
    e_R &= \frac{p_t^{(V)}-c_{RV}p_t^{(R)}}{s_{RV}} .
\end{aligned}
\label{eq:semantic_angle}
\end{equation}
Here $e_R$ is the unit direction of the shortest great-circle arc from $p_t^{(R)}$ to $p_t^{(V)}$, and $\theta_{RV}$ is its exact geodesic angle rather than an arbitrary Euclidean residual norm. The logarithmic map is therefore
\begin{equation}
    \xi_R
    =
    \operatorname{Log}_{p_t^{(R)}}(p_t^{(V)})
    =
    \theta_{RV}e_R
    \in T_{p_t^{(R)}}\mathbb S^{d-1}.
    \label{eq:semantic_log_map}
\end{equation}
When $p_t^{(V)}=p_t^{(R)}$, we define $\xi_R=0$ by the continuous limit of Eq.~\eqref{eq:semantic_log_map}; the local $\mathcal S$-space interventions considered here exclude the antipodal case. By construction, $\langle\xi_R,p_t^{(R)}\rangle_F=0$ and $\|\xi_R\|_F=\theta_{RV}$. Thus, the direction of $\xi_R$ specifies the local semantic trajectory and its norm has the explicit meaning of geodesic arc length. This avoids the ambiguity of treating an ambient residual magnitude as semantic strength.

\paragraph{Parallel transport to the real-image anchor.}
Although $\xi_R$ isolates the edit from the reconstruction gap, it belongs to the tangent space at the reconstructed reference $p_t^{(R)}$, while the Navigator should act at the real source point $p_t^{(N)}$. Directly reusing $\xi_R$ at $p_t^{(N)}$ is unreasonable because $\langle\xi_R,p_t^{(N)}\rangle_F\neq0$. We therefore parallel-transport $\xi_R$ along the shortest geodesic from $p_t^{(R)}$ to the real-image anchor $p_t^{(N)}$ and denote the resulting tangent vector by $\xi_t^{*}$. On the unit sphere, it has the closed form
\begin{equation}
    \xi_t^{*}
    =
    \xi_R
    -
    \frac{\langle\xi_R,p_t^{(N)}\rangle_F}
         {1+\langle p_t^{(R)},p_t^{(N)}\rangle_F}
    (p_t^{(R)}+p_t^{(N)}).
\label{eq:tangent_target}
\end{equation}
Because the locally paired source and reconstruction states are non-antipodal, the denominator in Eq.~\eqref{eq:tangent_target} is nonzero. The transported vector satisfies $\langle\xi_t^{*},p_t^{(N)}\rangle_F=0$ and $\|\xi_t^{*}\|_F=\|\xi_R\|_F=\theta_{RV}$. A target defined directly at the source, $\operatorname{Log}_{p_t^{(N)}}(p_t^{(V)})$, would mix the intended semantic displacement with the StyleGAN reconstruction gap: even when $\Delta s=0$ and $p_t^{(V)}=p_t^{(R)}$, it remains nonzero because $p_t^{(R)}\neq p_t^{(N)}$. In contrast, our target first isolates the edit as $\operatorname{Log}_{p_t^{(R)}}(p_t^{(V)})$ and then transports it to $p_t^{(N)}$. Consequently, $\Delta s=0\Rightarrow\xi_t^*=0$, while nonzero edits remain tangent at the real-image state and retain their semantic geodesic length.

\subsection{Navigator Learning and Stable Inference}\label{sec:method_navi_and_optim}
\paragraph{Tangent-field learning.}
The Cross-space Navigator learns a shared conditional mapping to the geometric target constructed above. Conditioned on the factorized $\mathcal S$-space displacement, timestep, and unedited source feature, it predicts $u_t=\mathcal M(\Delta s,t,h_t^{(N)})$.
The Navigator learns a shared, factor-conditioned tangent field in diffusion $h$-space without requiring an edited endpoint at inference. Whereas Asyrp learns a separate additive residual predictor for each prescribed semantic direction~\cite{kwon2022diffusionmodels}, $\mathcal M$ is shared across $\mathcal S$-space factors and directly predicts the required local trajectory from $\Delta s$. At inference, it requires neither an edited reference nor a direction-specific model.

The target $\xi_t^*$ is strictly tangent at $p_t^{(N)}$, whereas the raw network output need not be. We train only $\mathcal M$ while keeping the inversion encoder, StyleGAN generator, and diffusion U-Net frozen:
\begin{equation}
    \mathcal L_{\mathrm{tan}}
    =
    \mathbb E_{\substack{
        t\sim\mathcal{U}\{t_{\mathrm{edit}},\ldots,T\},\;
        \epsilon\sim\mathcal N(\mathbf 0,\mathbf I)
    }}
    \left[
        \left\|
        u_t-\xi_t^{*}
        \right\|_F^2
    \right].
    \label{eq:tan_loss}
\end{equation}
Sampling procedures for the semantic factor $k$ and intervention strength $\alpha$ are detailed in Supplementary Appendix~A.
Because $\xi_t^*\perp p_t^{(N)}$, this loss admits the orthogonal decomposition
\begin{equation}
    \|u_t-\xi_t^*\|_F^2
    =
    \|\Pi_{p_t^{(N)}}(u_t)-\xi_t^*\|_F^2
    +
    \langle u_t,p_t^{(N)}\rangle_F^2,
\label{eq:tangent_loss_decomposition}
\end{equation}
where
\begin{equation}
    \Pi_{p_t^{(N)}}(u_t)
    =
    u_t-\langle u_t,p_t^{(N)}\rangle_Fp_t^{(N)}.
    \label{eq:tangent_projection}
\end{equation}
Hence, the objective aligns the semantic tangent and penalizes radial leakage. At inference, we explicitly apply Eq.~\eqref{eq:tangent_projection} and denote $\widehat\xi_t=\Pi_{p_t^{(N)}}(u_t)$, ensuring that the final displacement remains tangent despite Navigator errors.

\paragraph{Geodesic realization.}
We realize the predicted tangent through the exponential map:
\begin{equation}
    \widetilde p_t
    =
    \cos(\|\widehat\xi_t\|_F)p_t^{(N)}
    +
    \sin(\|\widehat\xi_t\|_F)
    \frac{\widehat\xi_t}{\|\widehat\xi_t\|_F}.
\label{eq:navigator_inference}
\end{equation}
We then restore the source radius as $\widetilde h_t=r_N\widetilde p_t$, thereby preserving the source bottleneck scale and its correlation with the source skip-feature statistics and avoiding the norm-induced artifacts of unconstrained residual injection~\cite{jeong2024trainingfreecontent}. When $\widehat\xi_t=0$, as expected for the identity edit $\Delta s=0$, the normalized direction in Eq.~\eqref{eq:navigator_inference} is undefined. We therefore use the continuous limit $\operatorname{Exp}_{p_t^{(N)}}(0)=p_t^{(N)}$, so the bottleneck remains unchanged. For $\widehat\xi_t\neq0$, $\widehat\xi_t/\|\widehat\xi_t\|_F$ is a unit tangent orthogonal to $p_t^{(N)}$. Together, these cases ensure $\|\widetilde p_t\|_F=1$, $\|\widetilde h_t\|_F=\|h_t^{(N)}\|_F$, and $d_{\mathbb S}(p_t^{(N)},\widetilde p_t)=\|\widehat\xi_t\|_F$. Therefore, GeoNaviDiff neither introduces radial norm drift nor follows a straight chord through the shell interior. No additional geometric scaling is introduced: $\Delta s=\alpha d_k$ specifies the requested semantic factor and strength, while $\mathcal M$ adapts the corresponding tangent displacement to the source state and timestep.
\begin{table*}[tp]
    \centering
    \caption{Comparison with existing methods across VAE-, GAN-, and diffusion-based generative models. $\dagger$ denotes our baseline, \textbf{bold} indicates the best result, and ``--'' denotes not measured. Asyrp lacks fixed latent codes for PPL and DCI.}
    \label{tab:comparisons}
    \scriptsize
    \resizebox{\textwidth}{!}{
    \begin{tabular}{@{}lcccccccccccc@{}}
    \toprule
    \multirow{2}{*}{Method} 
    & \multicolumn{4}{c}{FFHQ} 
    & \multicolumn{4}{c}{AFHQ-v2} 
    & \multicolumn{4}{c}{Stanford Cars} \\ 
    \cmidrule(lr){2-5} \cmidrule(lr){6-9} \cmidrule(l){10-13}
    & FID$\downarrow$ & PPL$\downarrow$ & OFlow$\downarrow$ & DCI$\uparrow$
    & FID$\downarrow$ & PPL$\downarrow$ & OFlow$\downarrow$ & DCI$\uparrow$
    & FID$\downarrow$ & PPL$\downarrow$ & OFlow$\downarrow$ & DCI$\uparrow$ \\ 
    \midrule
    $\beta$-TCVAE~\cite{chen2018isolating} 
    & 96.50$\pm$2.40 & 128.85 & 0.12 & 0.28
    & 72.00$\pm$2.17 & 132.69 & 0.13 & 0.32
    & 92.78$\pm$2.35 & 133.54 & 0.13 & 0.31 \\

    FactorVAE~\cite{kim2018disentangling} 
    & 104.80$\pm$2.88 & 118.37 & 0.11 & 0.36
    & 84.50$\pm$2.60 & 122.56 & 0.14 & 0.35
    & 115.48$\pm$3.52 & 120.78 & 0.12 & 0.40 \\ 
    \midrule

    e4e~\cite{tov2021designing} 
    & 75.74$\pm$1.49 & 130.45 & 0.30 & 0.36
    & 26.65$\pm$0.55 & 142.69 & 0.23 & 0.31
    & 84.08$\pm$1.93 & 147.28 & 0.34 & 0.42 \\

    SFE~\cite{bobkov2024stylefeatureeditor} 
    & 76.42$\pm$1.35 & 114.88 & 0.17 & 0.54
    & 31.84$\pm$1.12 & 125.71 & 0.19 & 0.49
    & 79.21$\pm$1.70 & 122.75 & 0.27 & 0.47 \\

    NaviGAN~\cite{cherepkov2021navigating} 
    & 76.31$\pm$1.24 & 94.33 & 0.10 & 0.60
    & 49.36$\pm$1.09 & 88.95 & 0.12 & 0.57
    & 53.60$\pm$0.92 & 93.57 & 0.13 & 0.62 \\ 
    \midrule

    Asyrp$\dagger$~\cite{kwon2022diffusionmodels} 
    & 53.91$\pm$0.95 & -- & 0.18 & --
    & 20.90$\pm$2.42 & -- & 0.15 & --
    & 28.52$\pm$1.68 & -- & 0.19 & -- \\

    EncDiff~\cite{yang2024diffusioncrossattention}
    & 168.9$\pm$1.23 & 145.67 & 0.17 & 0.61
    & 91.2$\pm$1.08 & 135.64 & 0.18 & 0.53
    & 168.31$\pm$1.25 & 178.95 & 0.21 & 0.42 \\

    InfoDiffusion~\cite{wang2023infodiffusion}
    & 154.1$\pm$0.87 & 123.59 & 0.16 & 0.59
    & 87.5$\pm$1.35 & 138.30 & 0.14 & 0.48
    & 144.9$\pm$0.89 & 158.14 & 0.17 & 0.60 \\

    PreciseControl~\cite{parihar2024precisecontrolenhancing} 
    & 52.64$\pm$0.88 & 113.56 & 0.11 & 0.48
    & 21.72$\pm$0.60 & 140.66 & 0.12 & 0.51
    & 24.60$\pm$0.64 & 122.31 & 0.13 & 0.43 \\

    DisDiff~\cite{yang2023disdiffunsupervised}
    & 160.31$\pm$0.87 & 145.92 & 0.17 & 0.56
    & 172.24$\pm$0.71 & 145.32 & 0.15 & 0.43
    & 150.38$\pm$0.75 & 175.72 & 0.20 & 0.49 \\

    \textbf{GeoNaviDiff(Ours)} 
    & \textbf{26.78$\pm$0.39} & \textbf{42.36} & \textbf{0.08} & \textbf{0.76}
    & \textbf{13.66$\pm$0.25} & \textbf{68.55} & \textbf{0.07} & \textbf{0.73}
    & \textbf{18.92$\pm$0.38} & \textbf{71.39} & \textbf{0.06} & \textbf{0.69} \\ 
    \bottomrule
    \end{tabular}}
\end{table*}

\paragraph{Source-anchored rollout.}
During training, we sample timesteps independently and always condition $\mathcal M$ on an unedited source feature. During a reverse process, however, recursively querying $\mathcal M$ with the edited feature would expose it to unseen states. Earlier prediction errors would alter subsequent queries, and repeated residuals would cause semantic, identity, and structural drift, as verified by the ablation in Sec.~\ref{sec:ablation}. To remove this train--inference mismatch, GeoNaviDiff uses a source-anchored rollout (Alg.~1, Appendix~B). After DDIM inversion, we evolve an unedited source trajectory $\{X_t^{(\mathrm{src})}\}$ in parallel with the edited trajectory $\{X_t^{(\mathrm{edit})}\}$. At each step, the source branch extracts $h_t^{(\mathrm{src})}=H_\psi(X_t^{(\mathrm{src})},t)$ and normalizes it as $p_t^{(\mathrm{src})}=h_t^{(\mathrm{src})}/\|h_t^{(\mathrm{src})}\|_F$. It then queries $u_t^{(\mathrm{src})}=\mathcal M(\Delta s,t,h_t^{(\mathrm{src})})$ and constructs the absolute endpoint
\begin{equation}
    \widetilde h_t^{(\mathrm{edit}\mid\mathrm{src})}
    =
    \|h_t^{(\mathrm{src})}\|_F
    \operatorname{Exp}_{p_t^{(\mathrm{src})}}
    \left(
    \Pi_{p_t^{(\mathrm{src})}}(u_t^{(\mathrm{src})})
    \right).
\label{eq:source_anchored_endpoint}
\end{equation}
We inject the source-anchored endpoint into the decoder processing $X_t^{(\mathrm{edit})}$ and apply InjectFusion's latent calibration~\cite{jeong2024trainingfreecontent} to preserve bottleneck--state compatibility, yielding $X_{t-1}^{(\mathrm{edit})}$. This update avoids recursive residual accumulation on edited features, limiting train--inference mismatch and cumulative navigation error.

\begin{figure}[h!]
    \centering
    \includegraphics[width=\columnwidth]{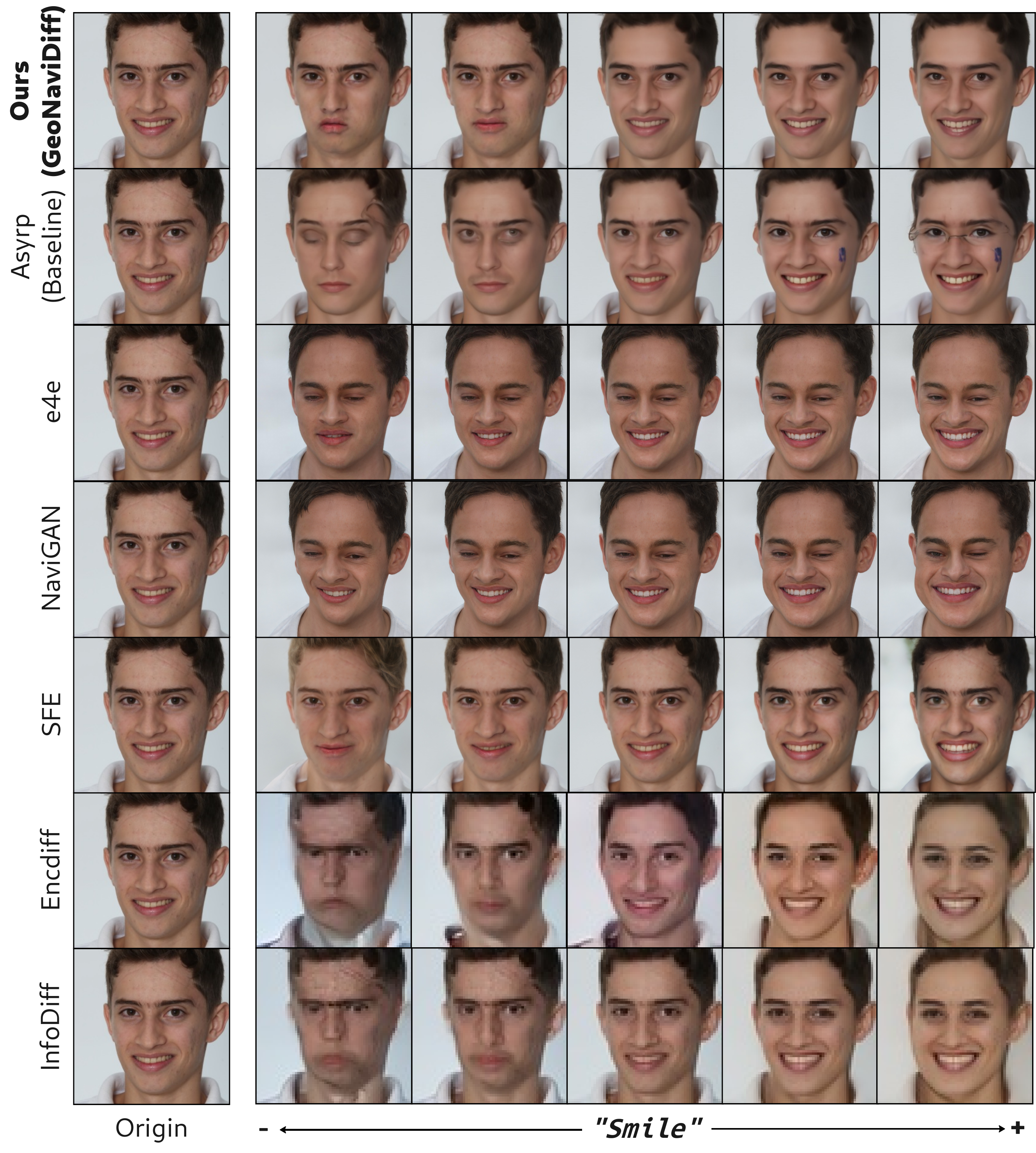}
    \caption{Qualitative comparison on smile manipulation across traversal levels.}
    \label{fig:comparison}
\end{figure}

\section{Experiments}

\subsection{Experimental Setup}\label{sec:experimental_setup}
\textbf{Datasets.}
We evaluate GeoNaviDiff on FFHQ~\cite{karras2019style}, CelebA-HQ~\cite{karras2017progressive}, AFHQ-v2~\cite{choi2020starganv2}, and Stanford Cars~\cite{krause20133d}, spanning human faces, animal faces, and vehicles. Details on preprocessing, data splits, hardware, and computational cost are provided in Supplementary Appendix~G.

\textbf{Metrics.}
For all four datasets, we report FID~\cite{heusel2017ganstrained} for distributional fidelity, Perceptual Path Length (PPL)~\cite{karras2019style} for perceptual smoothness along semantic traversals, OFlow~\cite{jin2024closedloop} for spatial editing leakage, and the disentanglement component of DCI ~\cite{eastwood2018framework} to measure how selectively individual representation dimensions encode semantic factors. The PPL computation protocol and dataset-specific DCI factor definitions are provided in Supplementary Appendix~A. Lower FID, PPL, and OFlow are better, whereas higher DCI is better.

\begin{figure*}[tp]
    \centering
    \includegraphics[width=0.85\textwidth]{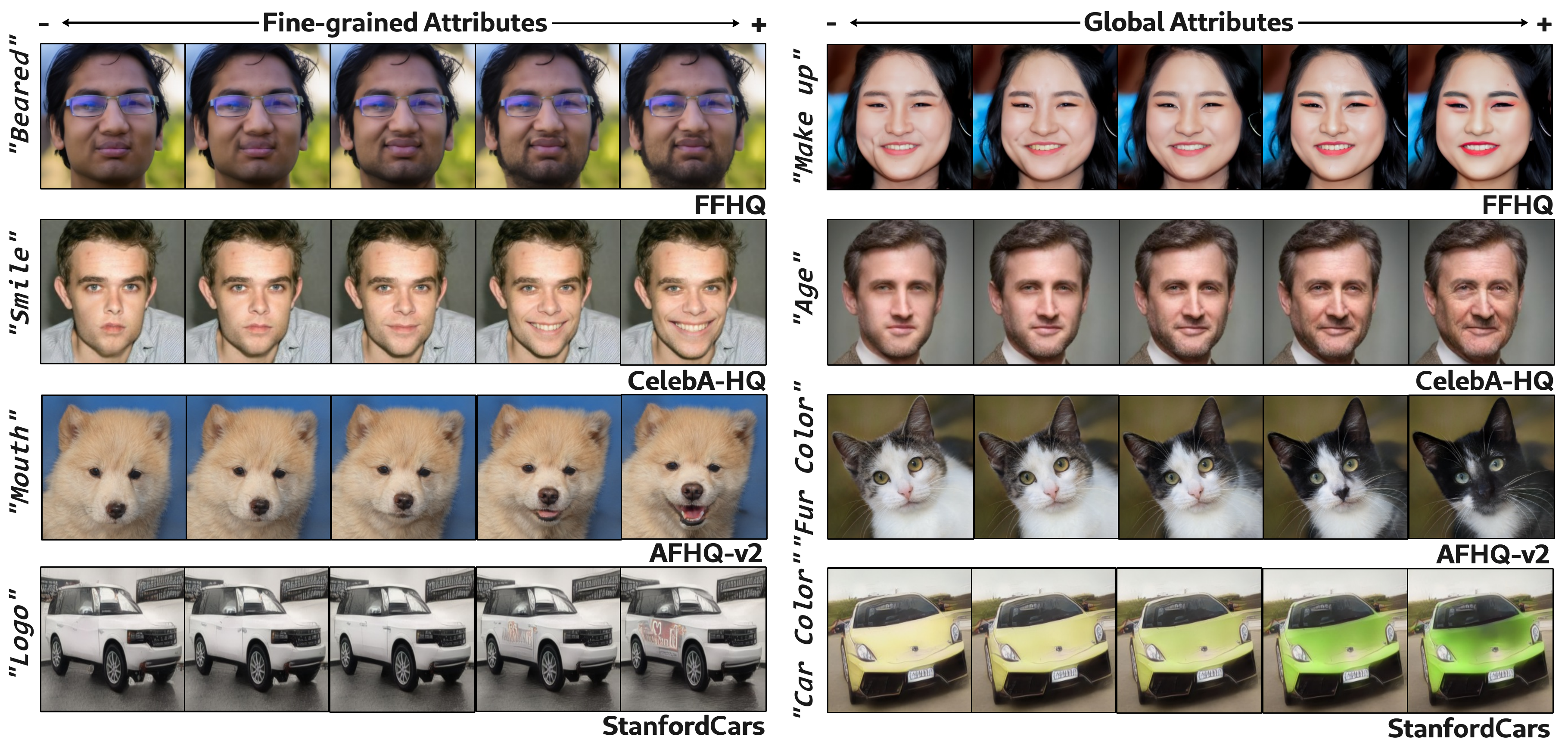}
    \caption{Qualitative traversals of GeoNaviDiff across datasets and semantic factors.}
    \label{fig:qualitative_traversal}
\end{figure*}
\begin{figure*}[t]
    \centering
    \includegraphics[width=0.85\textwidth]{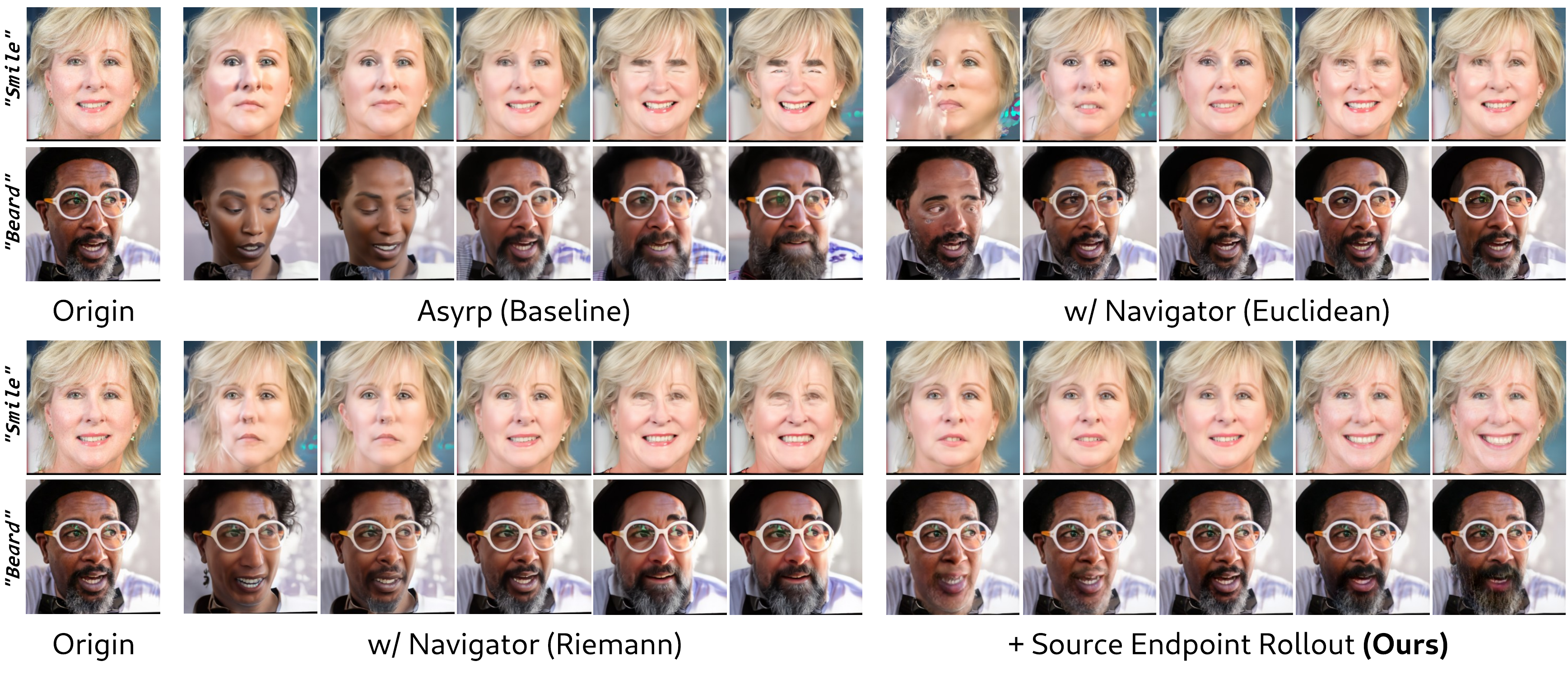}
    \caption{Stepwise ablation of cross-space navigation, Riemannian realization, and source-anchored rollout.}
    \label{fig:ablation}
\end{figure*}

\subsection{Comparison with Existing Methods}

Table~\ref{tab:comparisons} compares GeoNaviDiff with representative VAE-, GAN-, and diffusion-based methods, and Figure~\ref{fig:comparison} provides a visual comparison on smile manipulation. GeoNaviDiff achieves the best results on all reported metrics across the evaluated datasets: lower FID and PPL indicate better image fidelity and smoother semantic traversals, while lower OFlow and higher DCI show reduced non-target leakage and stronger factor-wise disentanglement. The visual comparison is consistent with these trends, where GeoNaviDiff produces more monotonic semantic changes while better preserving identity and unrelated facial attributes. These improvements support our central design: $\mathcal S$ supplies factor-selective semantic controls, while diffusion $h$-space provides the high-fidelity realization space; Riemannian navigation and source-anchored inference further reduce feature-geometry mismatch and accumulated non-target drift. More qualitative comparisons are provided in Supplementary Appendix~E.

\subsection{Qualitative results and Source-Space Analysis}\label{sec:qualitative}
Figure~\ref{fig:qualitative_traversal} shows GeoNaviDiff traversals across FFHQ, CelebA-HQ, AFHQ-v2, and Stanford Cars, covering both localized and globally coordinated semantic factors. Across traversal levels, GeoNaviDiff produces smooth semantic changes while preserving identity, structure, and unrelated content, indicating that the controls discovered in $\mathcal S$ remain interpretable after being transferred to diffusion $h$-space. Additional GeoNaviDiff traversals across datasets and semantic factors are provided in Supplementary Appendix~F.
A quantitative comparison of semantic source spaces and transformation strategies is provided in Supplementary Appendix~C.

\subsection{Ablation Study}\label{sec:ablation}
Figure~\ref{fig:ablation} presents a stepwise ablation using matched source images and semantic strengths. All GeoNaviDiff variants share the same Navigator architecture and inference settings. Starting from Asyrp's direction-specific Euclidean residual, the variants successively introduce the $\mathcal S$-conditioned Navigator, Riemannian realization, and source-anchored rollout. The progression shows that the Navigator improves target-attribute consistency, Riemannian realization better preserves identity and non-target appearance, and source anchoring yields the most stable trajectories at stronger edits by suppressing cumulative drift.The corresponding quantitative ablation results are reported in Supplementary Appendix~D.


\section{Conclusion}
We presented GeoNaviDiff, an unsupervised cross-space framework that maps interpretable $\mathcal S$-space directions to local Riemannian trajectories in diffusion $h$-space, decoupling semantic representation from image realization. Across multiple datasets, GeoNaviDiff improves representation disentanglement, image fidelity, and trajectory smoothness while preserving non-target content, supporting semantic-basis transfer as an alternative to disentangling diffusion features from scratch.

\bibliography{bibtex/WANDM_paper_exports,bibtex/sec_1_append,bibtex/method_refs}

\end{document}